\documentclass[runningheads]{llncs}
\usepackage{graphicx}
\usepackage{tikz}
\usepackage{comment}
\usepackage{amsmath,amssymb} % define this before the line numbering.
\usepackage{color}

\usepackage[accsupp]{axessibility}  % Improves PDF readability for those with disabilities.

\usepackage[pagebackref=true,breaklinks,colorlinks,citecolor=blue,linkcolor=blue,urlcolor=blue]{hyperref}

\usepackage{xcolor}         % colors
\usepackage{amsmath}
\usepackage{graphicx}
\usepackage{multirow}
\usepackage{color, colortbl}
\definecolor{Gray}{gray}{0.90}
\usepackage{times}
\usepackage{epsfig}
\usepackage{graphicx}
\usepackage{amsmath}
\usepackage{amssymb}
\usepackage{array}
\renewcommand\arraystretch{1.0}
\usepackage{booktabs}
\usepackage{xspace}
\usepackage{wrapfig}
\usepackage[rightcaption]{sidecap}
\usepackage{tabu}
\usepackage{longtable}
\usepackage{bm}

\usepackage[font=small,skip=0pt]{caption}
\usepackage{etoolbox}
\BeforeBeginEnvironment{wraptable}{\setlength{\intextsep}{0pt}}
\BeforeBeginEnvironment{wrapfigure}{\setlength{\intextsep}{0pt}}

\newcommand\blfootnote[1]{%
  \begingroup
  \renewcommand\thefootnote{}\footnote{#1}%
  \addtocounter{footnote}{-1}%
  \endgroup
}

\makeatletter
\DeclareRobustCommand\onedot{\futurelet\@let@token\@onedot}
\def\@onedot{\ifx\@let@token.\else.\null\fi\xspace}

\def\eg{\emph{e.g}\onedot} 
\def\ie{\emph{i.e}\onedot}

\def\etal{\emph{et al}\onedot}
\definecolor{darkgrey}{RGB}{96,96,96}
\makeatother

\begin{document}
% \renewcommand\thelinenumber{\color[rgb]{0.2,0.5,0.8}\normalfont\sffamily\scriptsize\arabic{linenumber}\color[rgb]{0,0,0}}
% \renewcommand\makeLineNumber {\hss\thelinenumber\ % \hspace{6mm} \rlap{\hskip\textwidth\ % \hspace{6.5mm}\thelinenumber}}
% \linenumbers
\pagestyle{headings}
\mainmatter
\def\ECCVSubNumber{197}  % Insert your submission number here

\title{EdgeNeXt: Efficiently Amalgamated CNN-Transformer Architecture for Mobile Vision Applications}

% INITIAL SUBMISSION 
% \begin{comment}
\titlerunning{EdgeNeXt} 
\author{
Muhammad Maaz\inst{1}$^{*}$ \and
Abdelrahman Shaker\inst{1}$^{*}$ \and
Hisham Cholakkal\inst{1} \and
Salman Khan\inst{1,2} \and
Syed Waqas Zamir\inst{3} \and
Rao Muhammad Anwer\inst{1,4} \and
Fahad Shahbaz Khan\inst{1,5}
}
\institute{$^\text{1} $Mohamed bin Zayed University of AI \qquad $^\text{2} $Australian National University \qquad \\ $^\text{3} $Inception Institute of Artificial Intelligence (IIAI) \qquad $^\text{4} $Aalto University \qquad \\ $^\text{5} $Linköping University}
\authorrunning{Maaz et al.}
% \end{comment}
%******************

% CAMERA READY SUBMISSION
\begin{comment}
\titlerunning{Abbreviated paper title}
% If the paper title is too long for the running head, you can set
% an abbreviated paper title here
%
\author{First Author\inst{1}\orcidID{0000-1111-2222-3333} \and
Second Author\inst{2,3}\orcidID{1111-2222-3333-4444} \and
Third Author\inst{3}\orcidID{2222--3333-4444-5555}}
%
\authorrunning{F. Author et al.}
% First names are abbreviated in the running head.
% If there are more than two authors, 'et al.' is used.
%
\institute{Princeton University, Princeton NJ 08544, USA \and
Springer Heidelberg, Tiergartenstr. 17, 69121 Heidelberg, Germany
\email{lncs@springer.com}\\
\url{http://www.springer.com/gp/computer-science/lncs} \and
ABC Institute, Rupert-Karls-University Heidelberg, Heidelberg, Germany\\
\email{\{abc,lncs\}@uni-heidelberg.de}}
\end{comment}
%******************
\maketitle

\begin{abstract}
\blfootnote{\textsuperscript{*}Equal contribution}
In the pursuit of achieving ever-increasing accuracy, large and complex neural networks are usually developed. Such models demand high computational resources and therefore cannot be deployed on edge devices. It is of great interest to build resource-efficient general purpose networks due to their usefulness in several application areas. In this work, we strive to effectively combine the strengths of both CNN and Transformer models and propose a new efficient hybrid architecture EdgeNeXt. Specifically in EdgeNeXt, we introduce split depth-wise transpose attention (STDA) encoder that splits input tensors into multiple channel groups and utilizes depth-wise convolution along with self-attention across channel dimensions to implicitly increase the receptive field and encode multi-scale features. Our extensive experiments on classification, detection and segmentation tasks, reveal the merits of the proposed approach, outperforming state-of-the-art methods with comparatively lower compute requirements. Our EdgeNeXt model with 1.3M parameters achieves 71.2\% top-1 accuracy on ImageNet-1K, outperforming MobileViT with an absolute gain of 2.2\% with 28\% reduction in FLOPs. Further, our EdgeNeXt model with 5.6M parameters achieves 79.4\% top-1 accuracy on ImageNet-1K. The code and models are available at \url{https://t.ly/_Vu9}.
\end{abstract}

\section{Introduction}
Convolutional neural networks (CNNs) and the recently introduced vision transformers (ViTs) have significantly advanced the state-of-the-art in several mainstream computer vision tasks, including object recognition, detection and segmentation \cite{schmidhuber2015deep,khan2021transformers}. The general trend is to make the network architectures more deeper and sophisticated in the pursuit of ever-increasing accuracy. While striving for higher accuracy, most existing CNN and ViT-based architectures ignore the aspect of computational efficiency (\ie, model size and speed) which is crucial to operating on resource-constrained devices such as mobile platforms. In many real-world applications \eg, robotics and self-driving cars, the recognition process is desired to be both accurate and have low latency on resource-constrained mobile platforms.

Most existing approaches typically utilize carefully designed efficient variants of convolutions to achieve a trade-off between speed and accuracy on resource-constrained mobile platforms \cite{squeezenet,ShuffleNetV2,MobileNetV2}. 
Other than these approaches, few existing works~\cite{MobileNetV3,mnasnet} employ hardware-aware neural architecture search (NAS) to build low latency accurate models for mobile devices. While being easy to train and efficient in encoding local image details, these aforementioned light-weight CNNs do not explicitly model global interactions between pixels.

\begin{wrapfigure}[23]{r}{0.5\textwidth}
  \begin{center}
    \includegraphics[width=0.49\textwidth]{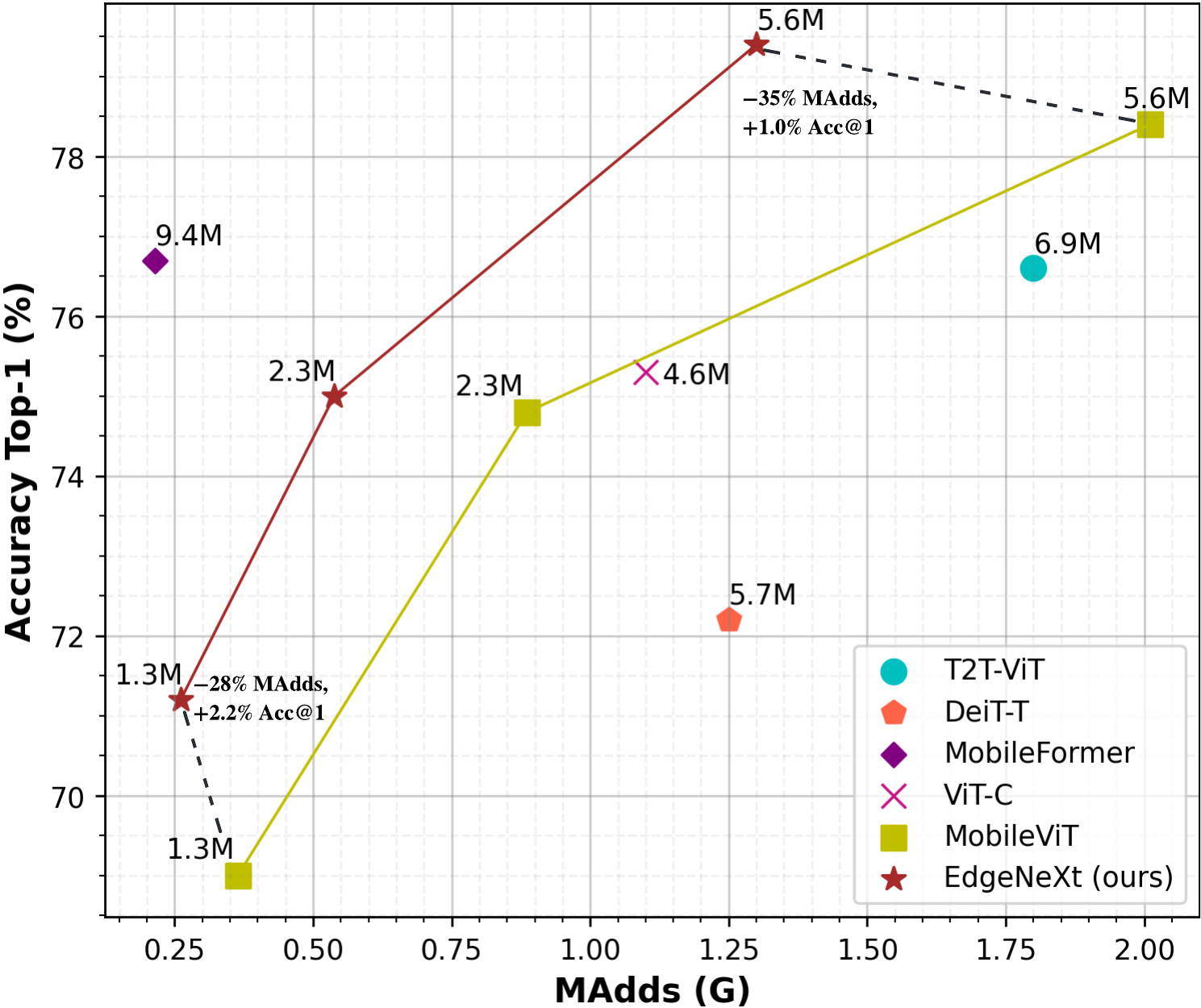}
  \end{center}
  \caption{Comparison of our proposed EdgeNeXt models with SOTA ViTs and hybrid architecture designs. The x-axis shows the multiplication-addition (MAdd) operations and y-axis displays the top-1 ImageNet-1K classification accuracy. The number of parameters are mentioned for each corresponding point in the graph. Our EdgeNeXt shows better compute (parameters and MAdds) versus accuracy trade-off compared to recent approaches.}
  \label{fig:tradeoff}
\end{wrapfigure}

The introduction of self-attention in vision transformers (ViTs)~\cite{ViTs} has made it possible to explicitly model this global interaction, however, this typically comes at the cost of slow inference because of the self-attention computation~\cite{liu2021swin}.
This becomes an important challenge for designing a lightweight ViT variant for mobile vision applications.

The majority of the existing works employ CNN-based designs in developing efficient models. However, the convolution operation in CNNs inherits two main limitations: First, it has local receptive field and thereby unable to model global context; Second, the learned weights are stationary at inference times, making CNNs inflexible to adapt to the input content. While both of these issues can be alleviated with Transformers, they are typically compute intensive. Few recent works \cite{EdgeFormer,MobileViT} have investigated designing lightweight architectures for mobile vision tasks by combining the strengths of CNNs and ViTs. However, these approaches mainly focus on optimizing the parameters and incur higher multiply-adds (MAdds) operations which restricts high-speed inference on mobile devices. The MAdds are %likely
higher since the complexity of the attention block is quadratic with respect to the input size ~\cite{MobileViT}. This becomes further problematic due to multiple attention blocks in the network architecture. Here, we argue that the model size, parameters, and MAdds are all desired to be small with respect to the resource-constrained devices when designing a unified mobile architecture that effectively combines the complementary advantages of CNNs and ViTs (see Fig.~\ref{fig:tradeoff}).

\noindent\textbf{Contributions.} We propose a new light-weight architecture, named \textit{EdgeNeXt}, that is efficient in terms of model size, parameters \textit{and} MAdds, while being superior in accuracy on mobile vision tasks. Specifically, we introduce split depth-wise transpose
attention (SDTA) encoder that effectively learns both local and global representations to address the issue of limited receptive fields in CNNs without increasing the number of parameters and MAdd operations. Our proposed architecture shows favorable performance in terms of both accuracy \textit{and} latency compared to state-of-the-art mobile networks on various tasks including image classification, object detection, and semantic segmentation. Our EdgeNeXt backbone with 5.6M parameters and 1.3G MAdds achieves 79.4\% top-1 ImageNet-1K classification accuracy which is superior to its recently introduced MobileViT counterpart~\cite{MobileViT}, while requiring 35\% less MAdds. For object detection and semantic segmentation tasks, the proposed EdgeNeXt achieves higher mAP and mIOU with fewer MAdds and a comparable number of parameters, compared to all the published lightweight models in literature.

\section{Related Work}
In recent years, designing lightweight hardware-efficient convolutional neural networks for mobile vision tasks has been well studied in literature. 
The current methods focus on designing efficient versions of convolutions for low-powered edge devices~\cite{squeezenet,MobileNet}.
Among these methods, MobileNet~\cite{MobileNet} is the most widely used architecture which employs depth-wise separable convolutions~\cite{DepthWiseConv}. On the other hand, ShuffleNet~\cite{ShuffleNet} uses channel shuffling and low-cost group convolutions. MobileNetV2~\cite{MobileNetV2} introduces inverted residual block with linear bottleneck, achieving promising performance on various vision tasks. ESPNetv2~\cite{espnetv2} utilizes depth-wise dilated convolutions to increase the receptive field of the network without increasing the network complexity. 
The hardware-aware neural architecture search (NAS) has also been explored to find a better trade-off between speed and accuracy on mobile devices~\cite{MobileNetV3,mnasnet}.
Although these CCNs are faster to train and infer on mobile devices, they lack global interaction between pixels which limits their accuracy.

Recently, Desovitskiy~\etal~\cite{ViTs} introduces a vision transformer architecture based on the self-attention mechanism~\cite{attention} for vision tasks. Their proposed architecture utilizes large-scale pre-training data (e.g. JFT-300M), extensive data augmentations, and a longer training schedule to achieve competitive performance.
Later, DeiT~\cite{DeiT} proposes to integrate distillation token in this architecture and only employ training on ImageNet-1K~\cite{imagenet} dataset. Since then, several variants of ViTs and hybrid architectures are proposed in the literature, adding image-specific inductive bias to ViTs for obtaining improved performance on different vision tasks \cite{BoTNet,convit,ViT-C,elsa,mvit2021}. 

ViT models achieve competitive results for several visual recognition tasks \cite{ViTs,liu2021swin}. However, it is difficult to deploy these models on resource-constrained edge devices because of the high computational cost of the multi-headed self-attention (MHA). There has been recent work on designing lightweight hybrid networks for mobile vision tasks that combine the advantages of CNNs and transformers. MobileFormer~\cite{MobileFormer} employs parallel branches of MobileNetV2~\cite{MobileNetV2} and ViTs~\cite{ViTs} with a bridge connecting both branches for local-global interaction. Mehta~\etal~\cite{MobileViT} consider transformers as convolution and propose a MobileViT block for local-global image context fusion. Their approach achieves superior performance on image classification surpassing previous light-weight CNNs and ViTs using a similar parameter budget. 

Although MobileViT~\cite{MobileViT} mainly focuses on optimizing parameters and latency, MHA is still the main efficiency bottleneck in this model, especially for the number of MAdds and the inference time on edge devices. The complexity of MHA in MobileViT is quadratic with respect to the input size, which is the main efficiency bottleneck given their existing nine attention blocks in MobileViT-S model. 
In this work, we strive to design a new light-weight architecture for mobile devices that is efficient in terms of both parameters and MAdds, while being superior in accuracy on mobile vision tasks.
Our proposed architecture, EdgeNeXt, is built on the recently introduced CNN method, ConvNeXt~\cite{ConvNeXt}, which modernizes the ResNet~\cite{resnet} architecture following the ViT design choices.
Within our EdgeNeXt, we introduce an SDTA block that combines depth-wise convolutions with adaptive kernel sizes along with transpose attention in an efficient manner, obtaining an optimal accuracy-speed trade-off. 

\section{EdgeNeXt}
%%%%%%%%%%%%%%%%%%%%%%%%%% BLOCK Diagram %%%%%%%%%%%%%%%%%%%%%%%%%%%

%%%%%%%%%%%%%%%%%%%%%%%%%% BLOCK Diagram %%%%%%%%%%%%%%%%%%%%%%%%%%%

The main objective of this work is to develop a lightweight hybrid design that effectively fuses the merits of ViTs and CNNs for low-powered edge devices. The computational overhead in ViTs (e.g., MobileViT~\cite{MobileViT}) is mainly due to the self-attention operation. In contrast to MobileViT, the attention block in our model has linear complexity with respect to the input spatial dimension of $\mathcal{O}(N d^2)$, where $N$ is the number of patches, and $d$ is the feature/channel dimension. The self-attention operation in our model is applied across channel dimensions instead of the spatial dimension.
Furthermore, we demonstrate that with a much lower number of attention blocks (3 versus 9 in MobileViT), we can surpass their performance mark.
In this way, the proposed framework can model global representations with a limited number of MAdds which is a fundamental criterion to ensure low-latency inference on edge devices.
To motivate our proposed architecture, we present two desirable properties.

\noindent\textbf{a) Encoding the global information efficiently.} The intrinsic characteristic of self-attention to learn global representations is crucial for vision tasks. To inherit this advantage efficiently, we use cross-covariance attention to incorporate the attention operation across the feature channel dimension instead of the spatial dimension within a relatively small number of network blocks. This reduces the complexity of the original self-attention operation from quadratic to linear in terms of number of tokens and implicitly encodes the global information effectively.

\noindent\textbf{b) Adaptive kernel sizes.} Large-kernel convolutions are known to be computationally expensive since the number of parameters and FLOPs quadratically increases as the kernel size grows. Although a larger kernel size is helpful to increase the receptive field, using such large kernels across the whole network hierarchy is expensive and sub-optimal. We propose an adaptive kernel sizes mechanism to reduce this complexity and capture different levels of features in the network. Inspired by the hierarchy of the CNNs, we use smaller kernels at the early stages, while larger kernels at the latter stages in the convolution encoder blocks. This design choice is optimal as early stages in CNN usually capture low-level features and smaller kernels are suitable for this purpose. However, in later stages of the network, large convolutional kernels are required to capture high-level features~\cite{zeiler2014visualizing}. We explain our architectural details next. 

\begin{figure}[!t]
    \centering
    \includegraphics[width=0.99\linewidth]{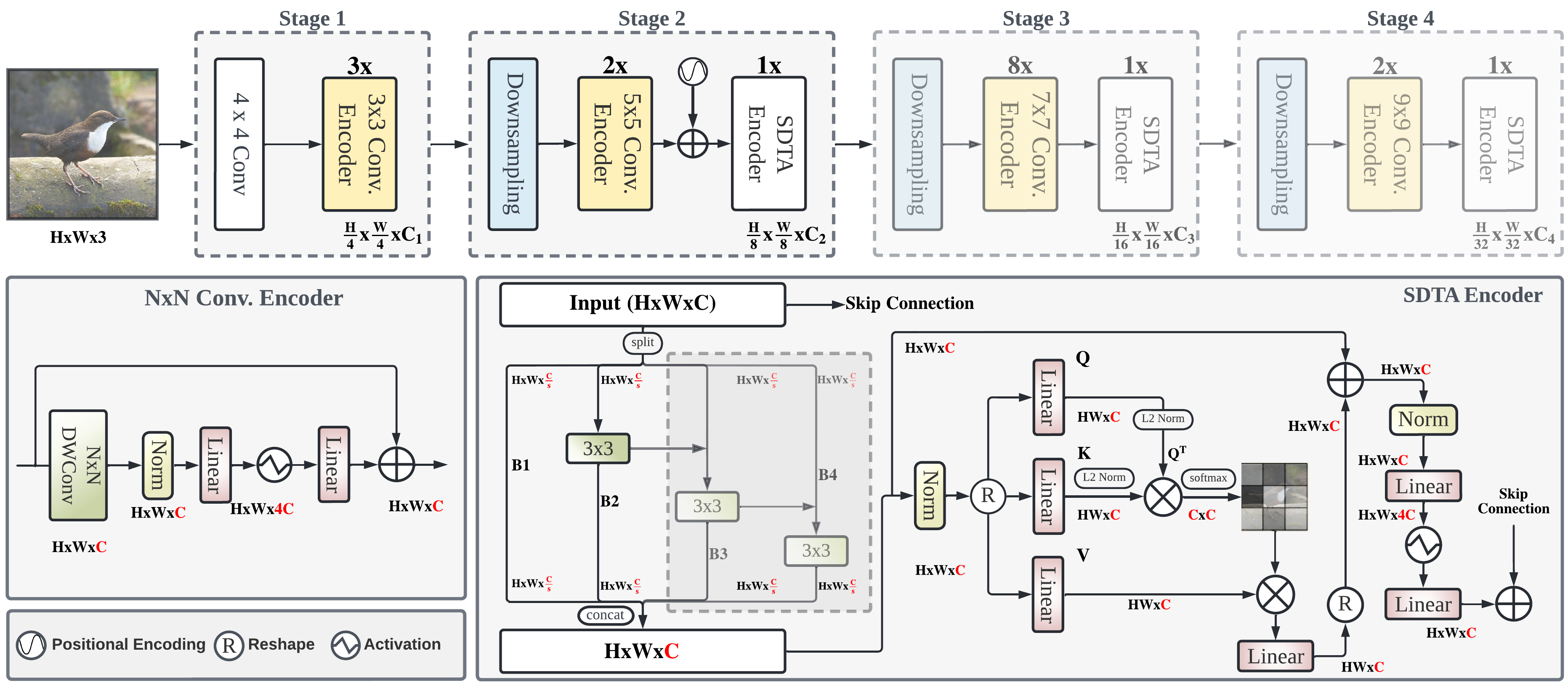}
    \caption{\textbf{Top Row:} The overall architecture of our framework is a stage-wise design. Here, the first stage downsamples the input image to $1/4^{th}$ resolution using $4\times4$ strided convolution followed by three $3\times3$ Convolution (Conv.) encoders. In stages 2-4, $2\times2$ strided convolutions are used for downsampling at the start, followed by $N$$\times$$N$ Convolution and the Split depth-wise Transpose Attention (SDTA) encoders. \textbf{Bottom Row:} We present the design of the Conv. encoder (Left) and the SDTA encoder (right). The Conv. encoder uses $N$$\times$$N$ depth-wise convolutions for spatial mixing followed by two pointwise convolutions for channel mixing. The SDTA Encoder splits the input tensor into $B$ channel groups and applies $3\times3$ depth-wise convolutions for multi-scale spatial mixing. The skip connections between branches increase the overall receptive field of the network. The branches $B3$ and $B4$ are progressively activated in stages 3 and 4, increasing the overall receptive field in the deeper layers of the network. Within the proposed SDTA, we utilize Transpose Attention followed by a light-weight MLP, that applies attention to feature channels and has linear complexity with respect to the input image.}
    \label{fig:Architecture}
\end{figure}

\noindent \textbf{Overall Architecture.} Fig.~\ref{fig:Architecture} illustrates an overview of the proposed EdgeNeXt architecture. The main ingredients are two-fold: \textbf{(1)} adaptive $N{\times}N$ Conv. encoder, and \textbf{(2)} split depth-wise transpose attention (SDTA) encoder. Our EdgeNeXt architecture builds on the design principles of ConvNeXt \cite{ConvNeXt} and extracts hierarchical features at four different scales across the four stages. The input image of size $H{\times}W{\times}3$ is passed through a patchify stem layer at the beginning of the network, implemented using a $4{\times}4$ non-overlapping convolution followed by a layer norm, which results in $\frac{H}{4}{\times}\frac{W}{4}{\times}C1$ feature maps. Then, the output is passed to 3$\times$3 Conv. encoder to extract local features. The second stage begins with a downsampling layer implemented using 2${\times}$2 strided convolution that reduces the spatial sizes by half and increases the channels, resulting in $\frac{H}{8}{\times}\frac{W}{8}{\times}C2$ feature maps, followed by two consecutive 5${\times}$5 Conv. encoders. Positional Encoding (PE) is also added before the SDTA block in the second stage only. 
We observe that PE is sensitive for dense prediction tasks (e.g., object detection and segmentation) as well as adding it in all stages increases the latency of the network.
Hence, we add it only once in the network to encode the spatial location information. The output feature maps are further passed to the third and fourth stages, to generate $\frac{H}{16}{\times}\frac{W}{16}{\times}C3$ and $\frac{H}{32}{\times}\frac{W}{32}{\times} C4$ dimensional features, respectively. 

\noindent \textbf{Convolution Encoder.} This block consists of depth-wise separable convolution with adaptive kernel sizes. We can define it by two separate layers: \textbf{(1)} depth-wise convolution with adaptive $N{\times}N$ kernels. We use $k$ = 3, 5, 7, and 9 for stages 1, 2, 3, and 4, respectively. Then, \textbf{(2)} two point-wise convolution layers are used to enrich the local representation alongside standard Layer Normalization~\cite{LN} (LN) and Gaussian Error Linear Unit~\cite{GELU} (GELU) activation for non-linear feature mapping. Finally, a skip connection is added to make information flow across the network hierarchy. This block is similar to the ConvNeXt block but the kernel sizes are dynamic and vary depending on the stage. We observe that adaptive kernel sizes in Conv. encoder perform better compared to static kernel sizes~(Table~\ref{Abl:tab6}). The Conv. encoder can be represented as follows:
\begin{equation}
    \bm{x}_{i+1} =  \bm{x}_{i} + Linear_G\big(Linear(LN(Dw(\bm{x}_{i})))\big),
\end{equation}
where $\bm{x}_{i}$ denotes the input feature maps of shape $H$${\times}$$W$${\times}$$C$, $Linear_G$ is a point-wise convolution layer followed by GELU, $Dw$ is $k$$\times$$k$ depth-wise convolution, $LN$ is a normalization layer, and $\bm{x}_{i+1}$ denotes the output feature maps of the Conv. encoder.

\noindent \textbf{SDTA Encoder.} There are two main components in the proposed split depth-wise transpose attention (SDTA)  encoder. The first component strives to learn an adaptive multi-scale feature representation by encoding various spatial levels within the input image and the second part implicitly encodes global image representations. The first part of our encoder is inspired by Res2Net~\cite{res2net} where we adopt a multi-scale processing approach by developing hierarchical representation into a single block. This makes the spatial receptive field of the output feature representation more flexible and adaptive. Different from Res2Net, the first block in our SDTA encoder does not use the $1 {\times1}$ pointwise convolution layers to ensure a lightweight network with a constrained number of parameters and MAdds.
Also, we use adaptive number of subsets per stage to allow effective and flexible feature encoding. In our STDA encoder, we split the input tensor $H$${\times}$$W$${\times}$$C$ into $s$ subsets, each subset is denoted by $\bm{x}_{i}$ and has the same spatial size with $C/s$ channels, where $i$ $\in$ \{1, 2, ..., $s$\} and $C$ is the number of channels. Each feature maps subset (except the first subset) is passed to $3{\times}3$ depth-wise convolution, denoted by $d_{i}$, and the output is denoted by $\bm{y}_{i}$. Also, the output of $d_{i-1}$, denoted by $\bm{y}_{i-1}$, is added to the feature subset $\bm{x}_{i}$, and then fed into $d_{i}$. The number of subsets $s$ is adaptive based on the stage number $t$, where $t$ $\in$ \{2, 3, 4\}. We can write $\bm{y}_{i}$ as follows: 
\begin{equation}
    \bm{y}_{i} =
    \begin{cases}
      \bm{x}_{i} & i=1;\\
      d_{i}(\bm{x}_{i}) & i=2, t=2;\\
      d_{i}(\bm{x}_{i}+\bm{y}_{i-1}) & 2<i \leq s,t.\\
    \end{cases} 
\end{equation}
Each depth-wise operation $d_{i}$, as shown in SDTA encoder in Fig.~\ref{fig:Architecture}, receives feature maps output from all previous splits \{$\bm{x}_{j}$, $j$ $\leq$ $i$\}.

As mentioned earlier, the overhead of the transformer self-attention layer is infeasible for vision tasks on edge-devices because it comes at the cost of higher MAdds and latency. To alleviate this issue and encode the global context efficiently, we use transposed query and key attention feature maps in our SDTA encoder~\cite{ali2021xcit}. This operation has a linear complexity by applying the dot-product operation of the MSA across channel dimensions instead of the spatial dimension, which allows us to compute cross-covariance across channels to generate attention feature maps that have implicit knowledge about the global representations. Given a normalized tensor $\bm{Y}$ of shape $H$${\times}$$W$${\times}$$C$, we compute query ($\bm{Q}$), key ($\bm{K}$), and value ($\bm{V}$) projections using three linear layers, yielding $\bm{Q}{=}\bm{W}^Q\bm{Y}$, $\bm{K}{=}\bm{W}^K\bm{Y}$, and $\bm{V}{=}\bm{W}^V\bm{Y}$ , with dimensions $HW$${\times}$$C$, where $\bm{W}^Q$,$\bm{W}^K$, and $\bm{W}^V$ are the projection weights for $\bm{Q}$, $\bm{K}$, and $\bm{V}$ respectively. Then, L2 norm is applied to $\bm{Q}$ and $\bm{K}$ before computing the cross-covariance attention as it stabilizes the training. Instead of applying the dot-product between $\bm{Q}$ and $\bm{K}^T$ along the spatial dimension i.e., ($HW$ ${\times}$ $C$) $\cdot$ ($C$ ${\times}$ $HW$), we apply the dot-product across the channel dimensions between $\bm{Q}^T$ and $\bm{K}$  i.e., ($C$$\times$$HW$) $\cdot$ ($HW$${\times}$$C$), producing $C$${\times}$$C$ softmax scaled attention score matrix. To get the final attention maps, we multiply the scores by $\bm{V}$ and sum them up. The transposed attention operation can be expressed as follows:
\begin{equation}
    \hat{\bm{X}} = Attention(\bm{Q},~\bm{K},~\bm{V}) + \bm{X},
    \label{eq:attention_1}
\end{equation}
\begin{equation}
   s.t., \; Attention(\bm{Q},~\bm{K},~\bm{V}) = \bm{V}\cdot \texttt{softmax}(\bm{Q}^T \cdot \bm{K})
    \label{eq:attention_2}
\end{equation}
where $\bm{X}$ is the input and $\hat{\bm{X}}$ is the output feature tensor. After that, two $1{\times}1$ pointwise convolution layers, LN and GELU activation are used to generate non-linear features. Table~\ref{METH:tab1} shows the sequence of Conv. and STDA encoders with the corresponding input size at each layer with more design details about extra-extra small, extra-small and small models.

\begin{table*}[!t]
\setlength{\tabcolsep}{6pt}
\renewcommand\arraystretch{1.0}
\caption{\textbf{EdgeNeXt Architectures.}~Description of the models' layers with respect to output size, kernel size, and output channels, repeated $n$ times, along with the models MAdds and parameters. The number of the output channels for small, extra-small, and extra-extra small models is chosen to match the number of parameters with the counterpart MobileViT model. We use adaptive kernel sizes in Conv.~Encoder to reduce the model complexity and capture different levels of features. Also, we pad the output size of the last stage to be able to apply the 9$\times$9 filter.}
\begin{center}
\begin{tabular}{l c c c c c c}
\toprule
\rowcolor{Gray} Layer     &  Output Size &  $n$ & Kernel & \multicolumn{3}{c}{Output Channels}\\
\midrule
 &   &   &    & \textbf{XXS} & \textbf{XS} & \textbf{S} \\
\midrule
Image & 256$\times$256 & 1 & - & - & - & - \\
\midrule
Stem & 64$\times$64 & 1  & 4$\times$4 & 24 & 32 & 48\\
Conv. Encoder & 64$\times$64 &  3 & 3$\times$3 & 24 & 32 & 48\\
\midrule
Downsampling & 32$\times$32 &  1 & 2$\times$2 & 48 & 64 & 96\\
Conv. Encoder & 32$\times$32 &  2 & 5$\times$5 & 48 & 64 & 96\\
STDA Encoder & 32$\times$32 &  1 & - & 48 & 64 & 96\\
\midrule
Downsampling & 16$\times$16 &  1 & 2$\times$2 & 88 & 100 & 160\\
Conv. Encoder & 16$\times$16 &  8 & 7$\times$7 & 88 & 100 & 160\\
STDA Encoder & 16$\times$16 &  1 & - &  88 & 100 & 160\\
\midrule
Downsampling & 8$\times$8 &  1 & 2$\times$2 & 168 & 192 & 304\\
Conv. Encoder & 8$\times$8 &  2 & 9$\times$9 & 168 & 192 & 304\\
STDA Encoder & 8$\times$8 &  1 & - &  168 & 192 & 304\\
\midrule
Global Average Pooling & 1$\times$1 &  1 & - & - & - & -\\
Linear & 1$\times$1 & 1 & - &  1000 & 1000 & 1000\\
\midrule
\textbf{Model MAdds}  &  &   &  & 0.3G & 0.5G & 1.3G\\
\textbf{Model Prameters}  &  &   &  & 1.3M & 2.3M & 5.6M\\
\bottomrule 
\end{tabular}
\label{METH:tab1}
\end{center}
\end{table*}

\section{Experiments}
In this section, we evaluate our EdgeNeXt model on ImageNet-1K classification, COCO object detection, and Pascal VOC segmentation benchmarks.

\subsection{Dataset}
We use ImageNet-1K~\cite{imagenet} dataset in all classification experiments. The dataset provides approximately 1.28M training and 50K validation images for 1000 categories. Following the literature~\cite{MobileNet,MobileViT}, we report top-1 accuracy on the validation set for all experiments.
For object detection, we use COCO~\cite{COCO} dataset which provides approximately 118k training and 5k validation images respectively. For segmentation, we use Pascal VOC 2012 dataset~\cite{voc} which provides almost 10k images with semantic segmentation masks. Following the standard practice as in~\cite{MobileViT}, we use extra data and annotations from~\cite{COCO} and~\cite{hariharan2011semantic} as well.

\subsection{Implementation Details}
We train our EdgeNeXt models at an input resolution of 256$\times$256 with an effective batch size of 4096. All the experiments are run for 300 epochs with AdamW~\cite{AdamW} optimizer, and with a learning rate and weight decay of 6e-3 and 0.05 respectively. We use cosine learning rate schedule~\cite{Cosine} with linear warmup for 20 epochs. The data augmentations used during training are Random Resized Crop (RRC), Horizontal Flip, and RandAugment~\cite{RandAugment}, where RandAugment is only used for the EdgeNeXt-S model. We also use multi-scale sampler~\cite{MobileViT} during training. Further stochastic depth~\cite{StochasticDepth} with a rate of 0.1 is used for EdgeNeXt-S model only. We use EMA~\cite{polyak1992acceleration} with a momentum of 0.9995 during training. For inference, the images are resized to 292${\times}$292 followed by a center crop at 256${\times}$256 resolution. We also train and report the accuracy of our EdgeNeXt-S model at 224${\times}$224 resolution for a fair comparison with previous methods. The classification experiments are run on eight A100 GPUs with an average training time of almost 30 hours for the EdgeNeXt-S model.

For detection and segmentation tasks, we finetune EdgeNeXt following similar settings as in~\cite{MobileViT} and report mean average precision (mAP) at IOU of
0.50-0.95 and mean intersection over union (mIOU) respectively. The experiments are run on four A100 GPUs with an average training time of $\sim$36 and $\sim$7 hours for detection and segmentation respectively.

We also report the latency of our models on NVIDIA Jetson Nano\footnote{https://developer.nvidia.com/embedded/jetson-nano-developer-kit} and NVIDIA A100 40GB GPU. For Jetson Nano, we convert all the models to TensorRT\footnote{https://github.com/NVIDIA/TensorRT} engines and perform inference in FP16 mode using a batch size of 1. For A100, similar to~\cite{ConvNeXt}, we use PyTorch v1.8.1 with a batch size of 256 to measure the latency.
%%%%%%%%%% Table 1 - Image Classification - Comparison with SoTA %%%%%%%%%% 
\begin{table*}[!t]
\small
\setlength{\tabcolsep}{6pt}
\renewcommand\arraystretch{1.0}
\caption{Comparisons of our proposed EdgeNeXt model with state-of-the-art lightweight fully convolutional, transformer-based and hybrid models on ImageNet-1K classification task. 
% EdgeNeXt-S$^*$ is trained using knowledge distillation following~\cite{usi}. 
Our model achieves a better trade-off between accuracy and compute (i.e., parameters and multiplication-addition (MAdds)).}
\begin{center}
% \resizebox{0.7\linewidth}{!}{
\begin{tabular}{l l l l r r c}
\toprule
\rowcolor{Gray} Frameworks  &   Models     &   Date  & Input &  Params$\downarrow$  & MAdds$\downarrow$ &   Top1$\uparrow$  \\
\midrule
                     &  MobileNetV2  &  CVPR2018  & 224$^2$ &  6.9M & 585M & 74.7 \\
ConvNets         &  ShuffleNetV2 &  ECCV2018  & 224$^2$ & 5.5M & 597M & 74.5 \\
                     &  MobileNetV3  &  ICCV2019  & 224$^2$ & 5.4M & 219M & 75.2 \\
\midrule
ViTs                 &  T2T-ViT                   &  ICCV2021  & 224$^2$ & 6.9M & 1.80G & 76.5 \\
                     &  DeiT-T                    &  ICML2021  & 224$^2$ & 5.7M & 1.25G & 72.2 \\
\midrule
                     &  MobileFormer              &  CoRR2021 & 224$^2$ & 9.4M & 214M & 76.7 \\
Hybrid               &  ViT-C                     &  NeurIPS2021 & 224$^2$ & 4.6M & 1.10G & 75.3 \\
                     &  CoaT-Lite-T               &  ICCV2021  & 224$^2$ & 5.7M & 1.60G  & 77.5 \\
                     &  MobileViT-S               &  ICLR2022  & 256$^2$ & 5.6M & 2.01G  & 78.4 \\
\cmidrule{2-7}
% Post-ConvNet         &  EdgeFormer-S              &  arXiv          & 256$^2$ & 5.0M & 1740M & 78.6 \\
\rowcolor{orange!6}  &  EdgeNeXt-S              &  Ours          & 224$^2$ & 5.6M & 965M & 78.8 \\
\rowcolor{orange!6}  &  EdgeNeXt-S              &  Ours          & 256$^2$ & 5.6M & 1.30G & 79.4 \\
% \cmidrule{2-7}
% \rowcolor{orange!6}  &  \textcolor{darkgrey}{EdgeNeXt-S$^*$}              &  \textcolor{darkgrey}{Ours}          & \textcolor{darkgrey}{256$^2$} & \textcolor{darkgrey}{5.6M} & \textcolor{darkgrey}{1.30G} & \textcolor{darkgrey}{81.1} \\
\bottomrule 
\end{tabular}
% }
\label{IN:tab1}
\end{center}
\end{table*}
%%%%%%%%%% Table 1 - Image Classification - Comparison with SoTA %%%%%%%%%%

% %%%%%%%%%% Table 1 - Image Classification - Comparison with SoTA %%%%%%%%%% 
% \begin{table*}[!t]
% \small
% \setlength{\tabcolsep}{6pt}
% \renewcommand\arraystretch{1.0}
% \caption{Comparisons of our proposed EdgeNeXt model with state-of-the-art lightweight fully convolutional, transformer-based and hybrid models on ImageNet-1K classification task. 
% % EdgeNeXt-S$^*$ is trained using knowledge distillation following~\cite{usi}. 
% Our model achieves a better trade-off between accuracy and compute (i.e., parameters and multiplication-addition (MAdds)).}
% \begin{center}
% % \resizebox{0.7\linewidth}{!}{
% \begin{tabular}{l l l r r c}
% \toprule
% \rowcolor{Gray} Models     &   Date  & Input &  Params$\downarrow$  & MAdds$\downarrow$ &   Top1$\uparrow$  \\
% \midrule
% EdgeNeXt-S              &  Ours          & 224$^2$ & 5.6M & 965M & 78.8 \\
% EdgeNeXt-S              &  Ours          & 256$^2$ & 5.6M & 1.30G & 79.4 \\
% % \cmidrule{2-7}
% \rowcolor{orange!6} {EdgeNeXt-S$^*$}              &  {Ours}          & {256$^2$} & \textcolor{darkgrey}{5.6M} & {1.30G} & {81.1} \\
% \bottomrule 
% \end{tabular}
% % }
% \label{IN:tab1}
% \end{center}
% \end{table*}
% %%%%%%%%%% Table 1 - Image Classification - Comparison with SoTA %%%%%%%%%%

%%%%% Table 2 - Image Classification - Comparison with MobileViT %%%%% 
\begin{table*}[!b]
\setlength{\tabcolsep}{4pt}
\renewcommand\arraystretch{1.0}
\caption{Comparison of different variants of EdgeNeXt with the counterpart models of MobileViT. The last two columns list the latency in $ms$ and $\mu s$ on NVIDIA Jetson Nano and A100 devices, respectively. Our EdgeNext models provide higher accuracy with lower latency for each model size, indicating the flexibility of our design to scale down to as few as 1.3M parameters.}
\begin{center}
\begin{tabular}{l l l r r c r r}
\toprule
\rowcolor{Gray}    Model     &   Date  & Input &  Params$\downarrow$  & MAdds$\downarrow$ &   Top1$\uparrow$ & Nano$\downarrow$ & A100$\downarrow$  \\
\midrule
MobileViT-XXS  &    &  & 1.3M & 364M  & 69.0 & 21.0 $ms$ & 216 $\mu s$ \\
MobileViT-XS  &  ICLR2022  & 256$^2$ & 2.3M & 886M  & 74.8 & 35.1 $ms$ & 423 $\mu s$ \\
MobileViT-S  &    &  & 5.6M & 2.01G  & 78.4 & 53.0 $ms$ & 559 $\mu s$ \\
\midrule
% \midrule
\rowcolor{orange!6}  EdgeNeXt-XXS   &     &  & 1.3M & 261M & 71.2 & 19.3 $ms$ & 142 $\mu s$ \\
\rowcolor{orange!6}  EdgeNeXt-XS   &  Ours   & 256$^2$ & 2.3M & 538M & 75.0 & 31.6 $ms$ &  227 $\mu s$ \\
\rowcolor{orange!6}  EdgeNeXt-S   &     &  & 5.6M & 1.30G & 79.4 & 48.8 $ms$ & 332 $\mu s$ \\
\bottomrule 
\end{tabular}
\label{IN:tab2}
\end{center}
\end{table*}
%%%%% Table 2 - Image Classification - Comparison with MobileViT %%%%% 

\subsection{Image Classification}
Table~\ref{IN:tab1} compares our proposed EdgeNeXt model with previous state-of-the-art fully convolutional (ConvNets), transformer-based (ViTs) and hybrid models. Overall, our model demonstrates better accuracy versus compute (parameters and MAdds) trade-off compared to all three categories of methods (see Fig.~\ref{fig:tradeoff}). 

\noindent\textbf{Comparison with ConvNets.} EdgeNeXt surpasses ligh-weight ConvNets by a formidable margin in terms of top-1 accuracy with similar parameters (Table~\ref{IN:tab1}). Normally, ConvNets have less MAdds compared to transformer and hybrid models because of no attention computation, however, they lack the global receptive field. For instance, EdgeNeXt-S has higher MAdds compared to MobileNetV2 \cite{MobileNetV2}, but it obtains 4.1\% gain in top-1 accuracy with less number of parameters. Also, our EdgeNeXt-S outperforms ShuffleNetV2 \cite{ShuffleNetV2} and MobileNetV3 \cite{MobileNetV3} by 4.3\% and 3.6\% respectively, with comparable number of parameters.

\noindent\textbf{Comparison with ViTs.} Our EdgeNeXt outperforms recent ViT variants on ImageNet1K dataset with fewer parameters and MAdds. For example, EdgeNeXt-S obtains 78.8\% top-1 accuracy, surpassing T2T-ViT \cite{T2T} and DeiT-T \cite{DeiT} by 2.3\% and 6.6\% absolute margins respectively.
 
\noindent\textbf{Comparison with Hybrid Models.} The proposed EdgeNeXt outperforms MobileFormer \cite{MobileFormer}, ViT-C \cite{ViT-C}, CoaT-Lite-T \cite{dai2021coatnet} with less number of parameters and fewer MAdds (Table~\ref{IN:tab1}). For a fair comparison with MobileViT~\cite{MobileViT}, we train our model at an input resolution of 256$\times$256 and show consistent gains for different models sizes (i.e., S, XS, and XXS) with fewer MAdds and faster inference on the edge devices (Table~\ref{IN:tab2}). For instance, our EdgeNeXt-XXS model achieves 71.2\% top-1 accuracy with only 1.3M parameters, surpassing the corresponding MobileViT version by 2.2\%. Finally, our EdgeNeXt-S model attains 79.4\% accuracy on ImageNet with only 5.6M parameters, a margin of 1.0\% as compared to the corresponding MobileViT-S model. This demonstrates the effectiveness and the generalization of our design.

Further, we also train our EdgeNeXt-S model using knowledge distillation following~\cite{usi} and achieves 81.1\% top-1 ImageNet accuracy.

\subsection{ImageNet-21K Pretraining}
To further explore the capacity of EdgeNeXt, we designed EdgeNeXt-B model with 18.5M parameters and 3.8MAdds and pretrain it on a subset of ImageNet-21K~\cite{imagenet} dataset followed by finetuning on standard ImageNet-1K dataset. ImageNet-21K (winter’21 release) contains around 13M images and 19K classes. We follow~\cite{ridnik2021imagenet} to preprocess the pretraining data by removing classes with fewer examples and split it into training and validation sets containing around 11M and 522K images respectively over 10,450 classes. We refer this dataset as ImageNet-21K-P. We strictly follow the training recipes of~\cite{ConvNeXt} for ImageNet-21K-P pretaining. Further, we initialize the ImageNet-21K-P training with ImageNet-1K pretrained model for faster convergence. Finally, we finetune ImageNet-21K model on ImageNet-1K for 30 epochs with a learning rate of 7.5e$^{-5}$ and an effective batch size of 512. The results are summarized in Table~\ref{IN21K}.

%%%%%%%%%%%%%%%%%%%%%%%%%%%%%%%% ImageNet-21K %%%%%%%%%%%%%%%%%%%%%%%%%%%%%%%%%
\begin{table*}[!t]
\setlength{\tabcolsep}{4pt}
\renewcommand\arraystretch{1.0}
\caption{Large-scale ImageNet-21K-P pretraining of EdgeNeXt-B model. Our model achieves better accuracy vs compute trade-off compared to SOTA ConvNeXt~\cite{ConvNeXt} and MobileViT-V2~\cite{mehta2022separable}.}
\begin{center}
\begin{tabular}{l c c r r c}
\toprule
\rowcolor{Gray}    Model & Pretraining  & Input &  Params$\downarrow$  & MAdds$\downarrow$ &   Top1$\uparrow$  \\
\midrule
ConvNeXt-T & None & 224$^2$ & 28.6M & 4.5G & 82.1 \\
ConvNeXt-T & ImageNet-21K & 224$^2$ & 28.6M & 4.5G & 82.9 \\
\midrule
MobileViT-V2 & None & 256$^2$ & 18.5M & 7.5G & 81.2 \\
MobileViT-V2 & ImageNet-21K-P & 256$^2$ & 18.5M & 7.5G & 82.4 \\
\midrule
\rowcolor{orange!6} EdgeNeXt-B & None & 256$^2$ & 18.5M & 3.8G & 82.5 \\
\rowcolor{orange!6} EdgeNeXt-B & ImageNet-21K-P & 256$^2$ & 18.5M & 3.8G & 83.3 \\
\bottomrule 
\end{tabular}
\label{IN21K}
\end{center}
\end{table*}
%%%%%%%%%%%%%%%%%%%%%%%%%%%%%%%% ImageNet-21K %%%%%%%%%%%%%%%%%%%%%%%%%%%%%%%%%

\subsection{Inference on Edge Devices}
We compute the inference time of our EdgeNeXt models on the NVIDIA Jetson Nano edge device and compare it with the state-of-the-art MobileViT~\cite{MobileViT} model (Table~\ref{IN:tab2}). All the models are converted to TensorRT engines and inference is performed in FP16 mode. Our model attains low latency on the edge device with similar parameters, fewer MAdds, and higher top-1 accuracy. Table~\ref{IN:tab2} also lists the inference time on A100 GPU for both MobileViT and EdgeNeXt models. It can be observed that our EdgeNeXt-XXS model is $\sim$34\% faster than the MobileViT-XSS model on A100 as compared to only $\sim$8\% faster on Jetson Nano, indicating that EdgeNeXt better utilizes the advanced hardware as compared to MobileViT. 

\subsection{Object Detection}
%%%%% Table 3 - COCO Object Detection - Comparison with SoTA %%%%% 
\begin{wraptable}[10]{r}{7cm}
\setlength{\tabcolsep}{3pt}
\resizebox{1.0\linewidth}{!}{
\begin{tabular}{l|r|r|c}
\toprule
\rowcolor{Gray}    Model  &  Params$\downarrow$  & MAdds$\downarrow$ &   mAP$\uparrow$ \\
\midrule
MobileNetV1 & 5.1M & 1.3G & 22.2 \\
MobileNetV2 & 4.3M & 800M & 22.1 \\
MobileNetV3 & 5.0M & 620M & 22.0 \\
\midrule
MobileViT-S & 5.7M & 3.4G & 27.7 \\
\rowcolor{orange!6} EdgeNeXt-S (ours) & 6.2M & 2.1G & 27.9 \\
\bottomrule 
\end{tabular}
}
\caption{Comparisons with SOTA on COCO object detection. EdgeNeXt improves over previous approaches.}
\label{OD:tab3}
\end{wraptable}
%%%%% Table 3 - COCO Object Detection - Comparison with SoTA %%%%% 

We use EdgeNeXt as a backbone in SSDLite and finetune the model on COCO 2017 dataset~\cite{COCO} at an input resolution of 320${\times}$320. The difference between SSD~\cite{liu2016ssd} and SSDLite is that the standard convolutions are replaced with separable convolutions in the SSD head. The results are reported in Table~\ref{OD:tab3}. EdgeNeXt consistently outperforms MobileNet backbones and gives competitive performance compared to MobileVit backbone. With less number of MAdds and a comparable number of parameters, EdgeNeXt achieves the highest 27.9 box AP, $\sim$38\% fewer MAdds than MobileViT.

%%%%% Table 4 - VOC Semantic Segmentation - Comparison with SoTA %%%%% 
\begin{wraptable}[9]{r}{7cm}
\caption{Comparisons with SOTA on VOC semantic segmentation. Our model provides reasonable gains.
}
\setlength{\tabcolsep}{3pt}
\resizebox{1.0\linewidth}{!}{
\begin{tabular}{l|r|r|c}
\toprule
\rowcolor{Gray}    Model &  Params$\downarrow$  & MAdds$\downarrow$ &   mIOU$\uparrow$  \\
\midrule
MobileNetV1 & 11.1M & 14.2G & 75.3 \\
MobileNetV2 & 4.5M & 5.8G & 75.7 \\
\midrule
MobileViT-S & 5.7M & 13.7G & 79.1 \\
\rowcolor{orange!6} EdgeNeXt-S (ours) & 6.5M & 8.7G & 80.2 \\
\bottomrule 
\end{tabular}
}
\label{tab:ablation_RKD_lvis}
\end{wraptable}
%%%%% Table 4 - VOC Semantic Segmentation - Comparison with SoTA %%%%% 

\subsection{Semantic Segmentation}
We use EdgeNeXt as backbone in DeepLabv3~\cite{chen2017rethinking} and finetune the model on Pascal VOC~\cite{voc} dataset at an input resolution of 512$\times$512. DeepLabv3 uses dilated convolution in cascade design along with spatial pyramid pooling to encode multi-scale features which are useful in encoding objects at multiple scales. Our model obtains 80.2 mIOU on the validation dataset, providing a 1.1 points gain over MobileViT with $\sim$36\% fewer MAdds.

\section{Ablations}
In this section, we ablate different design choices in our proposed EdgeNeXt model.

\noindent\textbf{SDTA encoder and adaptive kernel sizes.}
Table~\ref{Abl_sdta:tab5} illustrates the importance of SDTA encoders and adaptive kernel sizes in our proposed architecture. Replacing SDTA encoders with convolution encoders degrades the accuracy by 1.1\%, indicating the usefulness of SDTA encoders in our design. When we fix kernel size to 7 in all four stages of the network, it further reduces the accuracy by 0.4\%. Overall, our proposed design provides an optimal speed-accuracy trade-off.

We also ablate the contributions of SDTA components (e.g., adaptive branching and positional encoding) in Table~\ref{Abl_sdta:tab5}. Removing adaptive branching and positional encoding slightly decreases the accuracy.

%%%%% Table 5 - Ablations (SDTA Contribution & Components)  %%%%% 
\begin{SCtable}[\sidecaptionrelwidth][h!]
% \resizebox{0.6\linewidth}{!}{
\begin{tabular}{l|l|c|c}
\toprule
\rowcolor{Gray}  &  Model & Top1$\uparrow$ & Latency$\downarrow$  \\
\midrule
\rowcolor{orange!6} Base & EdgeNeXt-S & 79.4 & 332 $\mu$$s$ \\
\midrule
Different & w/o SDTA Encoders & 78.3 & 265 $\mu$$s$  \\
Components & + w/o Adaptive Kernels & 77.9 & 301 $\mu$$s$ \\
\midrule
SDTA & w/o Adaptive Branching  & 79.3 & 332 $\mu$$s$ \\
Components & + w/o PE & 79.2 & 301 $\mu$$s$ \\
\bottomrule 
\end{tabular}
% }
\caption{Ablation on different components of EdgeNeXt and SDTA encoder. The results show the benefits of SDTA encoders and adaptive kernels in our design. Further, adaptive branching and positional encoding (PE) are also required in SDTA module.}
\label{Abl_sdta:tab5}
\end{SCtable}
%%%%% Table 5 - Ablations (SDTA Contribution & Components) %%%%%

% \vspace{1em}
\noindent\textbf{Hybrid design.}
Table~\ref{Abl:tab6} ablates the different hybrid design choices for our EdgeNeXt model. Motivated from MetaFormer~\cite{yu2021metaformer}, we replace all convolutional modules in the last two stages with SDTA encoders. The results show superior performance when all blocks in the last two stages are SDTA blocks, but it increases the latency (row-2 vs 3). Our hybrid design where we propose to use an SDTA module as the last block in the last three stages provides an optimal speed-accuracy trade-off.

%%%%% Table 6 - Ablations (Hybrid Design)  %%%%%
% \vspace{-0.4cm}
\begin{SCtable}[\sidecaptionrelwidth][h!]
% \resizebox{0.65\linewidth}{!}{
\begin{tabular}{l|c|c}
\toprule
\rowcolor{Gray} Model Configuration & Top1$\uparrow$ & Latency$\downarrow$  \\
\midrule
1: Conv=[3, 3, 9, 0], SDTA=[0, 0, 0, 3] & 79.3 & 303 $\mu s$ \\
2: Conv=[3, 3, 0, 0], SDTA=[0, 0, 9, 3] & 79.7 & 393 $\mu s$ \\
\rowcolor{orange!6} 3: Conv=[3, 2, 8, 2], SDTA=[0, 1, 1, 1] & 79.4 & 332 $\mu s$ \\
\bottomrule 
\end{tabular}
% }
\caption{Ablation on the hybrid architecture. Using one SDTA encoder as the last block in the last three stages provides an optimal accuracy-latency trade-off.}
\label{Abl:tab6}
\end{SCtable}

%%%%% Table 6 - Ablations (Hybrid Design)  %%%%% 

Table~\ref{Abl:tab7} provides an ablation of the importance of using SDTA encoders at different stages of the network. It is noticable that progressively adding an SDTA encoder as the last block of the last three stages improves the accuracy with some loss in inference latency. However, in row 4, we obtain the best trade-off between accuracy and speed where the SDTA encoder is added as the last block in the last three stages of the network. Further, we notice that adding a global SDTA encoder to the first stage of the network is not helpful where the features are not much mature.

%%%%% Table 7 - Ablations (Hybrid Design)  %%%%% 
\begin{SCtable}[\sidecaptionrelwidth][h!]
% \resizebox{0.6\linewidth}{!}{
\begin{tabular}{l|c|c}
\toprule
\rowcolor{Gray} Model Configuration & Top1$\uparrow$ & Latency$\downarrow$  \\
\midrule
1: Conv=[3, 3, 9, 3], SDTA=[0, 0, 0, 0] & 78.3 & 265 $\mu s$ \\
2: Conv=[3, 3, 9, 2], SDTA=[0, 0, 0, 1] & 78.6 & 290 $\mu s$ \\
3: Conv=[3, 3, 8, 2], SDTA=[0, 0, 1, 1] & 79.1 & 310 $\mu s$ \\
\rowcolor{orange!6} 4: Conv=[3, 2, 8, 2], SDTA=[0, 1, 1, 1] & 79.4 & 332 $\mu s$ \\
5: Conv=[2, 2, 8, 2], SDTA=[1, 1, 1, 1] & 79.2 & 387 $\mu s$ \\
\bottomrule 
\end{tabular}
% }
\caption{Ablation on using SDTA encoder at different stages of the network. Including SDTA encoders in the last three stages improves performance, whereas a global SDTA encoder is not helpful in the first stage.}
\label{Abl:tab7}
\end{SCtable}
%%%%% Table 7 - Ablations (Hybrid Design)  %%%%% 

We also provide an ablation on using the SDTA module at the start of each stage versus at the end. Table~\ref{Abl:tab8} shows that using the global SDTA encoder at the end of each stage is more beneficial. This observation is consistent with the recent work~\cite{li2021improved}.

%%%%% Table 8 - Ablations (Hybrid Design)  %%%%% 
\begin{SCtable}[\sidecaptionrelwidth][h!]
% \resizebox{1.0\linewidth}{!}{
\begin{tabular}{l|c|c}
\toprule
\rowcolor{Gray} SDTA Configuration & Top1$\uparrow$ & Latency$\downarrow$  \\
\midrule
Start of Stage (SDTA=[0, 1, 1, 1]) & 79.0 & 332 $\mu s$ \\
\rowcolor{orange!6} End of Stage (SDTA=[0, 1, 1, 1]) & 79.4 & 332 $\mu s$ \\
\bottomrule 
\end{tabular}
% }
\caption{Ablation on using SDTA at the start and end of each stage in EdgeNeXt. The results show that it is generally beneficial to use SDTA at the end of each stage.}
\label{Abl:tab8}
\end{SCtable}
%%%%% Table 8 - Ablations (Hybrid Design)  %%%%% 

\begin{figure}[!h]
\centering
  \includegraphics[width=0.98\textwidth]{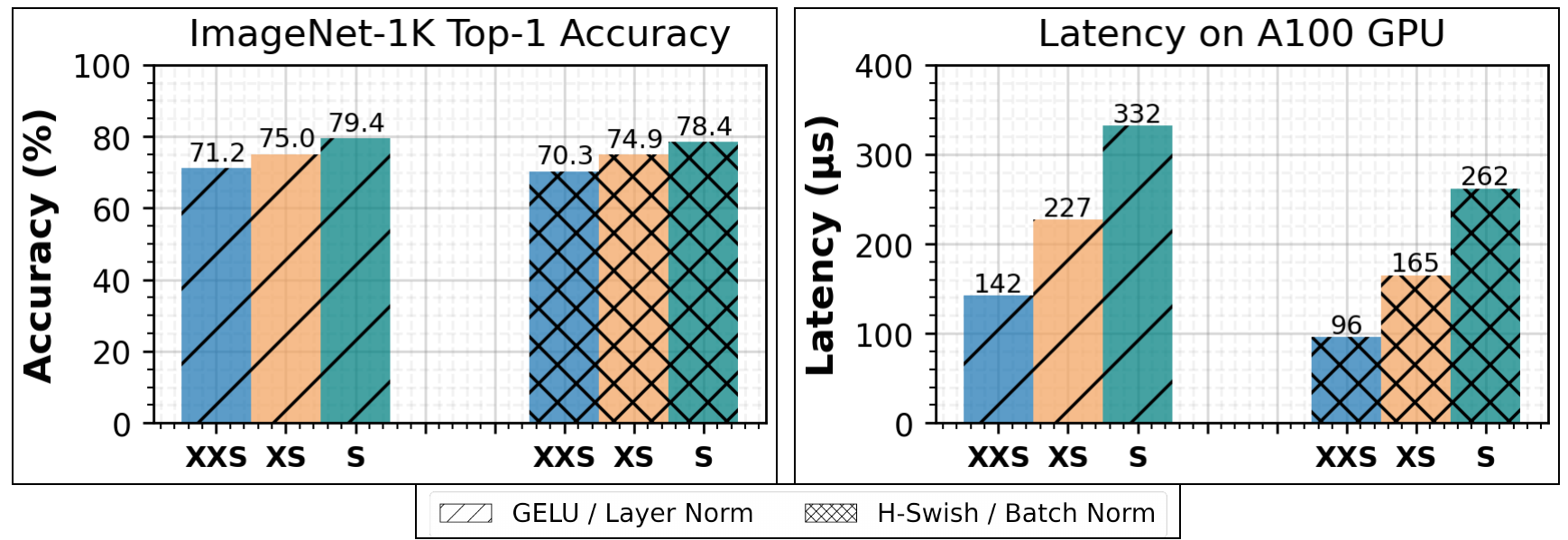}
  \caption{Ablation on the effect of using different activation functions and normalization layers on accuracy and latency of our network variants. Using Hard Swish activation and batch normalization instead of GELU and layer normalization significantly improves the latency at the cost of some loss in accuracy.}
  \label{Abl:fig3}
\end{figure}

% \vspace{-2em}
\noindent\textbf{Activation and normalization.}
EdgeNeXt uses GELU activation and layer normalization throughout the network. We found that the current PyTorch implementations of GELU and layer normalization are not optimal for high speed inference. To this end, we replace GELU with Hard-Swish and layer-norm with batch-norm and retrain our models. Fig.~\ref{Abl:fig3} indicates that it reduces the accuracy slightly, however, reduces the latency by a large margin.

\section{Qualitative Results}
\begin{figure}[!ht]
    \centering
    \includegraphics[width=0.92
    \linewidth]{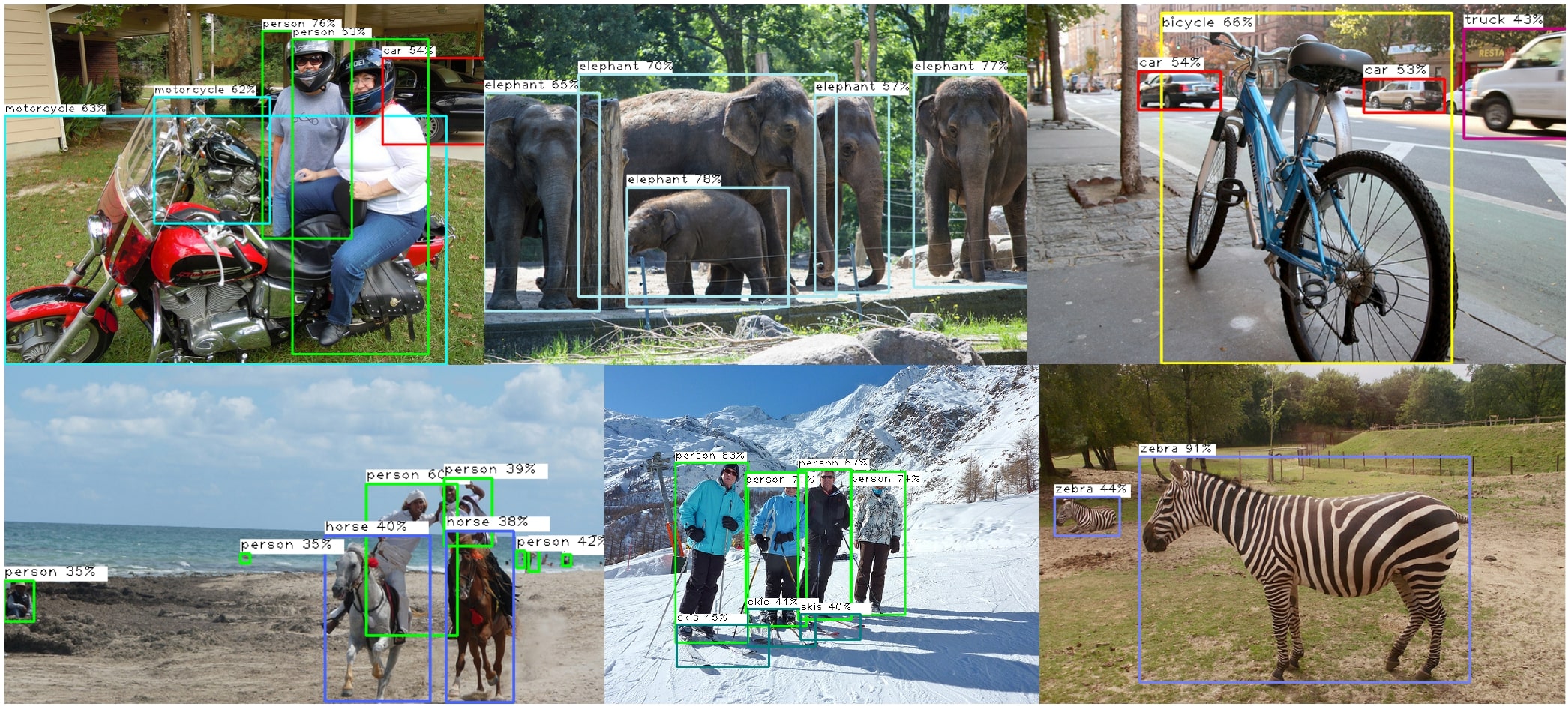}
    \caption{\small Qualitative results of our EdgeNeXt detection model on COCO validation dataset. The model is trained on COCO dataset with 80 detection classes. Our model can effectively localize and classify objects in diverse scenes.}
    \label{qualitative_detection}
\end{figure}

\begin{figure}[!ht]
    \centering
    \includegraphics[width=0.92\linewidth]{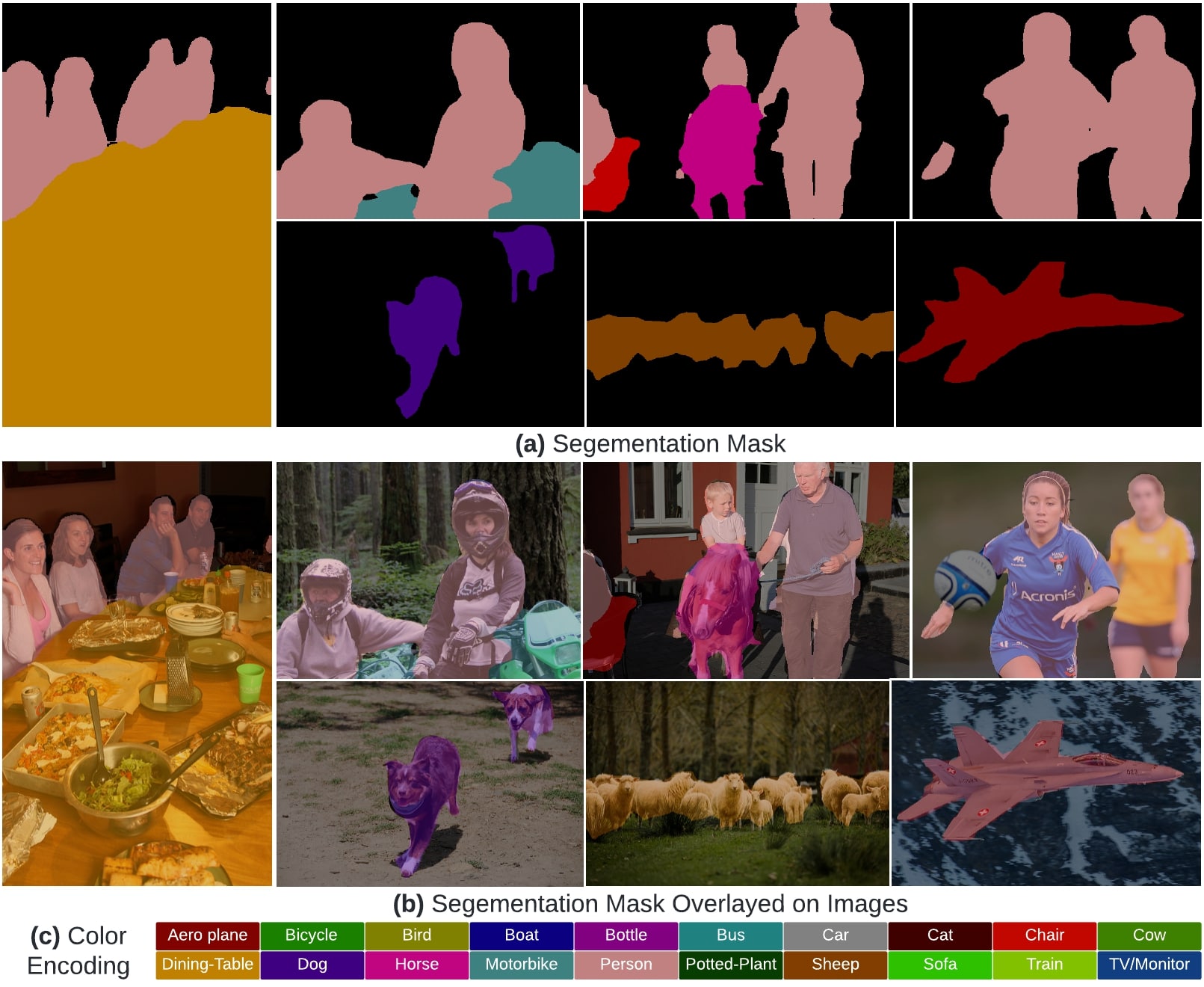}
    \caption{Qualitative results of our EdgeNeXt segmentation model on unseen COCO validation dataset. The model is trained on Pascal VOC dataset with 20 segmentation classes. {\color{blue}\textbf{(a)}} shows the predicted semantic segmentation mask where \lq black \rq color represents the background pixels. {\color{blue}\textbf{(b)}} displays the predicted masks on top of original images. {\color{blue}\textbf{(c)}} represents the color encodings for all Pascal VOC classes for the displayed segmentation masks. Our model provides high-quality segmentation masks on unseen COCO images.
    }
    \label{qualitative_segmentation}
\end{figure}

Figs.~\ref{qualitative_detection} and~\ref{qualitative_segmentation} shows the qualitative results of EdgeNeXt detection and segmentation models respectively. Our model can detect and segment objects in various views.

\section{Conclusion}
The success of the transformer models comes with a higher computational overhead compared to CNNs. Self-attention operation is the major contributor to this overhead, which makes vision transformers slow on the edge devices compared to CNN-based mobile architectures. In this paper, we introduce a hybrid design consisting of convolution and efficient self-attention based encoders to jointly model local and global information effectively, while being efficient in terms of both parameters and MAdds on vision tasks with superior performance compared to state-of-the-art methods. Our experimental results show promising performance for different variants of EdgeNeXt, which demonstrates the effectiveness and the generalization ability of the proposed model.

\clearpage
% ---- Bibliography ----
%
% BibTeX users should specify bibliography style 'splncs04'.
% References will then be sorted and formatted in the correct style.
%
\bibliographystyle{splncs04}
\bibliography{egbib}
\end{document}